\documentclass[runningheads]{llncs}
\usepackage[T1]{fontenc}
\usepackage{graphicx,verbatim}
\usepackage{xcolor, multirow, booktabs, adjustbox, amsmath, caption, color, colortbl}
\definecolor{lightblue}{HTML}{E6E5FD}
\usepackage[colorlinks=true, linkcolor=blue, urlcolor=blue, citecolor=blue]{hyperref}

\usepackage{color}
\begin{document}

\title{Retrieval-Augmented Anatomical Guidance for Text-to-CT Generation}

\author{Daniele Molino\inst{1} \and
Camillo Maria Caruso\inst{1} \and
Paolo Soda\inst{1,2} \and
Valerio Guarrasi\inst{1}
}
\authorrunning{D. Molino et al.}

\institute{Unit of Artificial Intelligence and Computer Systems, Department of Engineering, Università Campus Bio-Medico di Roma, Rome, Italy\\
\email{daniele.molino@unicampus.it, camillomaria.caruso@unicampus.it, valerio.guarrasi@unicampus.it, p.soda@unicampus.it}
\and
Department of Diagnostics and Intervention, Biomedical Engineering and Radiation Physics, Umeå University, Umeå, Sweden\\
\email{paolo.soda@umu.se}
}

\maketitle
\begin{abstract}
Text-conditioned generative models for volumetric medical imaging provide semantic control but lack explicit anatomical guidance, often resulting in outputs that are spatially ambiguous or anatomically inconsistent.
In contrast, structure-driven methods ensure strong anatomical consistency but typically assume access to ground-truth annotations, which are unavailable when the target image is to be synthesized.
We propose a retrieval-augmented approach for Text-to-CT generation that integrates semantic and anatomical information under a realistic inference setting.
Given a radiology report, our method retrieves a semantically related clinical case using a 3D vision-language encoder and leverages its associated anatomical annotation as a structural proxy.
This proxy is injected into a text-conditioned latent diffusion model via a ControlNet branch, providing coarse anatomical guidance while maintaining semantic flexibility.
Experiments on the CT-RATE dataset show that retrieval-augmented generation improves image fidelity and clinical consistency compared to text-only baselines, while additionally enabling explicit spatial controllability, a capability inherently absent in such approaches.
Further analysis highlights the importance of retrieval quality, with semantically aligned proxies yielding consistent gains across all evaluation axes.
This work introduces a principled and scalable mechanism to bridge semantic conditioning and anatomical plausibility in volumetric medical image synthesis.
Code is available at~\url{https://github.com/arco-group/RAGText2CT}. 

\end{abstract}

\section{Introduction}
Artificial Intelligence (AI) is increasingly integrated into medical imaging workflows~\cite{alyasseri2022review,litjens2016deep}, yet its large-scale deployment remains constrained by limited annotated data, privacy regulations, and high labeling cost~\cite{bansal2022systematic,guo2021multi,tajbakhsh2021guest}.
In this context, Generative AI offers a promising solution for synthesizing realistic medical data to support data augmentation, simulation, and privacy-aware learning~\cite{chlap2021review,frangi2018simulation,kazerouni2023diffusion,singh2021medical}.
Among imaging modalities, Computed Tomography (CT) plays a central clinical role by providing high-resolution volumetric representations~\cite{mazonakis2016computed}.
However, volumetric generation poses significant challenges in terms of computational scalability and global anatomical coherence.
A prominent research direction in this field is \emph{text-conditioned} generation, where radiology reports guide the synthesis process.
Radiological narratives provide high-level semantic descriptions derived from expert interpretation, enabling clinically aligned controllability.
Nevertheless, reports are inherently underspecified with respect to spatial structure: they describe pathological findings but do not encode explicit anatomical constraints and omit large portions of normal anatomy.
As a result, text-only conditioning may produce semantically plausible yet spatially ambiguous or anatomically inconsistent outputs.
Early text-to-CT approaches, such as GenerateCT~\cite{hamamci2024generatect} and MedSyn~\cite{xu2024medsyn}, rely on multi-stage pipelines combining low-resolution synthesis with super-resolution refinement, which may introduce inter-slice inconsistencies.
More recently, the growing interest in report-conditioned 3D CT generation, further catalyzed by the introduction of dedicated benchmarking efforts~\cite{challenge}, has led to diffusion-based architectures that can operate directly in compact volumetric latent spaces~\cite{molino2025text,amirrajab2025radiology}.
In parallel, \emph{structure-driven} methods, such as MAISI~\cite{guo2025maisi}, condition synthesis on explicit spatial inputs, e.g., segmentation masks.
While this paradigm enables precise anatomical control, it lacks semantic expressiveness and relies on ground-truth annotations at inference time, which are unavailable when the volume itself must be synthesized.

This contrast reveals a key limitation of existing methods: text-based conditioning provides semantic flexibility without explicit spatial control, whereas mask-based conditioning ensures anatomical precision at the cost of semantic richness and realistic inference assumptions.
Bridging these paradigms therefore requires leveraging anatomical information without observing or segmenting the target volume itself.
To this end, we here introduce a Retrieval-Augmented Generation (RAG)~\cite{chen2024benchmarking,khandelwal2020nearest,lewis2020retrieval} formulation extended to a multimodal volumetric setting that permits us to reinterpret anatomical structure as a \emph{retrievable latent proxy} rather than a direct conditioning input.
Indeed, rather than  assuming explicit access to structural annotations, we treat anatomical information as a latent source that can be approximated by retrieving relevant related examples from previously observed data.
In practice, given an input report, a pretrained 3D vision-language encoder retrieves semantically related clinical cases from a reference corpus.
The associated anatomical annotations are employed as coarse structural proxies that provide informative spatial constraints.
The retrieved proxy is then injected into a text-conditioned latent diffusion model via a ControlNet~\cite{zhang2023adding} branch, guiding synthesis toward anatomically coherent solutions while preserving semantic variability induced by the report.
Intuitively, the retrieved anatomy serves as a spatial scaffold rather than a precise template, allowing the generative process to adapt fine-grained structure while enforcing global anatomical coherence.

In summary, the contributions of this work are threefold:
\begin{itemize}
    \item We propose a retrieval-augmented framework for report-conditioned 3D CT synthesis that treats anatomical structure as a latent, retrievable proxy.
    \item We introduce a multimodal integration strategy that injects retrieved anatomical proxies into a text-conditioned latent diffusion model via ControlNet, enabling anatomical guidance without annotations at inference time.
    \item We provide extensive quantitative and qualitative evaluations of image fidelity, clinical consistency, and spatial controllability, and analyze the impact of retrieval quality on generation performance.
\end{itemize}

\section{Methods}
\begin{figure*}[t]
    \centering
    \includegraphics[width=0.84\textwidth]{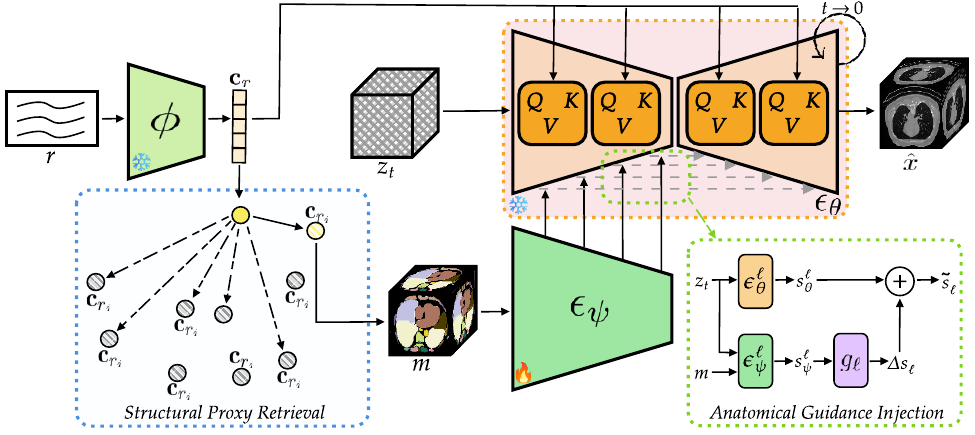}
    \caption{
    The proposed retrieval-augmented Text-to-CT generation framework.
    }
    \label{fig:method_overview}
\end{figure*}

\subsubsection{Problem Formulation}
\label{sec:problem_formulation}
We consider the task of generating a CT volume $\hat{x}$ conditioned on a radiology report $r$, assuming no access to the corresponding volume $x$ or structural annotations at inference time.
We therefore model anatomical structure as a latent source of information, named as {\em retrieved structural proxy} and denoted as $m$ in the following,  that can be approximated through retrieval. 
It is worth noting that in our formulation  $m$ is not assumed to match the target anatomy, but it acts as a plausible spatial scaffold  biasing the generative process toward anatomically coherent solutions.
Therefore, given the condition provided by the semantic ($r)$ and coarse anatomical ($m$) guidance,  the generation process is defined as:
\[
\hat{x} \sim p_\theta(x \mid r, m)
\]
where $p_\theta$ denotes a generative model. 
The overall retrieval-augmented generation framework is illustrated in Figure~\ref{fig:method_overview}.

\subsubsection{Text-to-CT Generative Backbone}
\label{sec:backbone}
We adopt a latent diffusion model for volumetric CT synthesis~\cite{molino2025text,rombach2022high} operating in a compressed latent space obtained through a variational autoencoder (VAE)~\cite{kingma2013auto}.
Diffusion is performed in the latent domain, and the final CT volume is reconstructed via the VAE decoder.
Text conditioning is implemented through embeddings extracted from a vision-language model based on the CLIP paradigm~\cite{radford2021learning}, extended with a 3D image encoder.
This formulation enables the alignment of radiology reports and CT volumes in a shared semantic embedding space, providing effective grounding of clinical language into volumetric representations~\cite{molino2025any,molino2025medcodi}.
The resulting report embeddings are used to guide generation according to the semantic content of the input text.
In this work, the diffusion backbone and text-conditioning mechanism are kept fixed, and we focus on augmenting the model with an anatomically informed conditioning pathway that does not alter the pretrained generative architecture.

\subsubsection{Retrieval-Augmented Structural Proxy}
\label{sec:rag_prior}
To approximate structural information in the absence of explicit anatomical inputs, we introduce a retrieval-based mechanism to obtain a structural proxy.
Given a report $r$, we retrieve a semantically related case from a reference corpus using a pretrained 3D vision-language encoder~\cite{molino2025text}, denoted as $\phi(\cdot)$.
From the retrieved case, we extract the associated anatomical annotation (e.g., a segmentation mask), which is used as the structural proxy $m$.
Formally, let $\mathbf{c}_r = \phi(r) \in \mathbf{R}^d$ be the embedding of the input report, given a reference set of reports $\{r_i\}_{i=1}^N$ with embeddings
$\{\mathbf{c}_{r_i} = \phi(r_i)\}_{i=1}^N$, the structural proxy is obtained as:
\[
m = \mathcal{M}\!\left(\arg\max_{i} \; \mathrm{sim}(\mathbf{c}_r, \mathbf{c}_{r_i})\right)
\]
where $\mathrm{sim}(\cdot,\cdot)$ denotes cosine similarity and
$\mathcal{M}(\cdot)$ is a deterministic operator that returns the anatomical annotation associated with the retrieved case.
Semantic similarity in the shared embedding space is hypothesized to correlate with coarse pathological and anatomical patterns, making the retrieved annotation a suitable, albeit noisy, structural proxy.
We retrieve the nearest neighbor to provide a deterministic and unambiguous signal and further analyze the effect of retrieval quality on generation performance in Section~\ref{sec:results}.

\subsubsection{Anatomical Guidance via ControlNet}
\label{sec:controlnet}
To integrate the retrieved structural proxy into the generative process, we inject anatomical guidance through a dedicated control branch that preserves the pretrained generative flow, thereby maintaining report-driven semantic variability while enforcing global anatomical consistency via the retrieved proxy.
To this end, we adopt a ControlNet-based conditioning mechanism~\cite{zhang2023adding}, where we introduce a trainable control branch $\epsilon_{\psi}$ alongside the frozen diffusion backbone $\epsilon_{\theta}$, mirroring its encoder architecture and weights.
Let $z_t$ denote the noisy latent at diffusion step $t$, $\mathbf{c}_r$ the report embedding, and $m$ the structural proxy.
During the forward pass of the frozen backbone, the encoder produces multi-scale skip features $\{s_\theta^{\ell}\}_{\ell=0}^{L}$ together with a bottleneck representation $b_{\theta}$,
\[
(\{s_\theta^{\ell}\}, b_{\theta}) = E_{\theta}(z_t, t, \mathbf{c}_r)
\]
which are used to predict the noise component in the decoder.
In parallel, the ControlNet branch processes the same noisy latent and semantic conditioning, augmented with the structural proxy $m$, yielding corresponding control features
\[
(\{s_\psi^{\ell}\}, b_{\psi}) = E_{\psi}(z_t, t, \mathbf{c}_r, m)
\]
The control features are not injected directly into the backbone activations.
Instead, we map $s^\ell_{\psi}$ and $b_{\psi}$ through zero-initialized convolutions to produce residual corrections
\[
\Delta s_\ell = \gamma\, g_\ell(s_\psi^{\ell}) \qquad
\Delta b = \gamma\, g_{\mathrm{mid}}(b_{\psi})
\]
where $g_\ell(\cdot)$ and $g_{\mathrm{mid}}(\cdot)$ denote zero-initialized projections and $\gamma$ is a scalar conditioning scale.
We then inject the residuals into the skip connections and the bottleneck of the frozen backbone,
\[
\tilde{s}_\ell = s_\theta^{\ell} + \Delta s_\ell \qquad
\tilde{b} = b_{\theta} + \Delta b
\]
and the final noise prediction is obtained by decoding the modulated features
\[
\hat{\epsilon} = D_{\theta}(\{\tilde{s}_\ell\}, \tilde{b}, t, \mathbf{c}_r)
\]
Zero-initialization ensures $\Delta s_\ell \approx 0$ and $\Delta b \approx 0$ at the start of training, recovering the original pretrained generator.
During training, the diffusion backbone and the vision-language encoder are kept frozen, and only the parameters of $E_\psi$ and the zero-initialized projection layers $g_\ell(\cdot)$ and $g_{\text{mid}}(\cdot)$ are optimized 
conditioning on ground-truth annotations via the rectified flow objective~\cite{liu2022flow}.
At inference time, the proxy $m$ is obtained exclusively through retrieval, enabling anatomically informed generation without access to explicit annotations.

\section{Experimental Configurations}
\label{sec:experimental}
\noindent\textbf{Datasets and Data Preparation}
Experiments are conducted on the CT-RATE dataset~\cite{hamamci2024developing}, comprising paired 3D chest CT volumes and radiology reports covering 18 thoracic pathologies.
We follow the official train/test split provided by the dataset, comprising 27,514 training volumes and 1,818 test volumes.
All CT volumes are resampled to $0.75 \times 0.75 \times 1.5$~mm and resized to $512 \times 512 \times 128$ voxels.
Intensity values are converted to Hounsfield Units, clipped to $[-1000, 1000]$, and normalized to $[0,1]$.
Anatomical segmentation masks extracted using TotalSegmentator~\cite{wasserthal2023totalsegmentator} are available for all volumes in the repository.
Retrieval is performed exclusively over the training set; test set reports are never included in the retrieval index and test set masks are never provided as conditioning input, ensuring no information leakage at evaluation time.

\noindent\textbf{Compared Methods}
We compare our approach against both text-conditioned and structure-conditioned generative models.
In the former group, we include GenerateCT~\cite{hamamci2024generatect} and MedSyn~\cite{xu2024medsyn}, representative multi-stage pipelines, as well as diffusion-based report-conditioned models Text-to-CT~\cite{molino2025text} and Report2CT~\cite{amirrajab2025radiology}.
These baselines enable isolating the effect of anatomical guidance.
While in the latter, we include MAISI~\cite{guo2025maisi}, which conditions directly on ground-truth segmentation masks.
Since MAISI does not incorporate textual conditioning, it is not directly comparable in terms of semantic controllability.
Instead, it serves as an upper bound, illustrating the level of anatomical consistency achievable when exact annotations are available at inference time.
Additionally, to assess the role of retrieval quality, we perform an ablation study evaluating three strategies for selecting the structural proxy: semantically nearest, semantically farthest, and random retrieval in the shared vision-language embedding space.
All configurations share the same generative backbone, differing exclusively in the proxy selection criterion at inference time.
This ablation isolates the impact of semantic alignment between the report and the retrieved structural proxy.

\noindent\textbf{Evaluation Metrics}
We evaluate generative performance along three complementary axes: image fidelity, clinical consistency, and spatial controllability.
\emph{Image Fidelity.} Visual realism is assessed using FID score~\cite{heusel2017gans}.
We report both slice-based FID computed on axial, coronal, and sagittal views using a 2D medical backbone, and volumetric FID using a 3D medical backbone.

\noindent\emph{Clinical Consistency.} Clinical plausibility is evaluated using CT-Net~\cite{draelos2021machine}, a 3D CNN trained on real CTs and kept frozen during evaluation.
Classification performance on synthetic CTs measures alignment with the conditioning reports.

\noindent\emph{Spatial Controllability.} Spatial controllability is evaluated using Dice score and $95^{th}$ percentile Hausdorff Distance (HD95).
Segmentation masks are predicted from generated volumes using TotalSegmentator~\cite{wasserthal2023totalsegmentator} and compared against the corresponding reference masks.
Dice quantifies volumetric overlap, while HD95 captures boundary-level geometric deviations, providing a complementary assessment of structural adherence to the provided mask.

\noindent\textbf{Implementation Details}
All components are implemented in PyTorch.
Control features are injected at all downsampling and bottleneck layers.
The ControlNet conditioning scale is fixed to 1.0 for all experiments, selected via a sweep over $[0.5, 2.0]$ balancing structural adherence and semantic flexibility, while text conditioning is applied using classifier-free guidance with a guidance scale of 5.0.
Sampling is performed using rectified flow with 30 steps.
The ControlNet branch is trained for 50 epochs using AdamW with learning rate of $1\times10^{-5}$ and weight decay $1\times10^{-2}$.
Training uses a batch size of 4, distributed across two NVIDIA A100 GPUs, and is performed with mixed-precision arithmetic.

\section{Results}
\label{sec:results}
This section presents a comprehensive evaluation of the proposed retrieval-augmented Text-to-CT approach, with the goal of isolating the contribution of the retrieved structural proxy across three complementary evaluation axes: image fidelity, clinical consistency, and spatial controllability. 
* in tables denotes statistically significantly inferior performance compared to the proposed method according to a paired Wilcoxon signed-rank test across 5 independent runs with different random seeds ($p < 0.05$).

\begin{table}[ht]
\caption{FID scores computed using 2.5D and 3D feature extractors.}
\label{tab:fid}
\centering
\scalebox{0.9}{
\setlength{\tabcolsep}{6pt}
\begin{tabular}{l|cccc|c}
\toprule
\multirow{2}{*}{\textbf{Method}} & \multicolumn{4}{c|}{\textbf{FID 2.5D ↓}} & \multirow{2}{*}{\textbf{FID 3D ↓}} \\
                                  & \textbf{Axial} & \textbf{Coronal} & \textbf{Sagittal} & \textbf{Avg.} & \\
\midrule
GenerateCT~\cite{hamamci2024generatect} & 6.701* & 11.793* & 9.694* & 9.396* & 0.166* \\
MedSyn~\cite{xu2024medsyn} & 8.789* & 8.900* & 8.717* & 8.802* & 0.169* \\
Report2CT~\cite{amirrajab2025radiology} & 1.345* & 2.294* & 1.855* & 1.775* & 0.043* \\
Text-to-CT~\cite{molino2025text} & 0.500* & 0.519* & 0.561* & 0.527* & 0.015* \\
MAISI~\cite{guo2025maisi}  & 0.462* & 0.568* & 0.503* & 0.511* & 0.012* \\
\midrule
RAG-Farthest & 0.298* & 0.399* & 0.340* & 0.346* & 0.010* \\
RAG-Random & 0.271 & 0.343* & 0.341* & 0.311* & 0.005 \\
\rowcolor{lightblue} RAG-Nearest & 0.275 & 0.317 & 0.318 & 0.303 & 0.004 \\
\bottomrule
\end{tabular}
}
\end{table}

\noindent\textbf{Image Fidelity}
Table~\ref{tab:fid} reports FID results in both 2.5D and 3D settings.
Across all configurations, introducing a structural proxy improves image fidelity with respect to text-only generation.
Notably, RAG variants also outperform MAISI; we attribute this to the absence of semantic conditioning in MAISI, which produces anatomically consistent volumes that are nonetheless misaligned with the semantic distribution of test reports, which FID directly penalizes.
All retrieval-augmented variants achieve lower FID scores than competing methods, indicating improved global anatomical coherence.
Semantically nearest retrieval yields the most stable improvements, while random and farthest retrieval result in slightly degraded FID scores. 
This suggests that retrieval quality has a modest effect on low-level appearance statistics, while playing a more prominent role in higher-level semantic and anatomical alignment.

\begin{table}[!t]
  \centering
  \begin{minipage}{0.6\textwidth}
    \centering
        \caption{Clinical consistency evaluation of text-conditioned models using CT-Net.}
    \label{tab:ctnet}
    \resizebox{\linewidth}{!}{
    \begin{tabular}{l|cc|cc}
\toprule
\multirow{2}{*}{\textbf{Model}} 
& \multicolumn{2}{c|}{\textbf{AUC} $\uparrow$} 
& \multicolumn{2}{c}{\textbf{Precision} $\uparrow$} \\
& \textbf{Macro} & \textbf{Weighted} 
& \textbf{Macro} & \textbf{Weighted} \\
\midrule
GenerateCT~\cite{hamamci2024generatect} & 0.581* & 0.568* & 0.285* & 0.361* \\
MedSyn~\cite{xu2024medsyn}             & 0.560* & 0.547* & 0.282* & 0.362* \\
Report2CT~\cite{amirrajab2025radiology} & 0.720* & 0.705* & 0.436* & 0.508* \\
Text-to-CT~\cite{molino2025text}       & 0.745* & 0.731* & 0.477* & 0.549* \\
\midrule
RAG-Farthest       & 0.712* & 0.716* & 0.429* & 0.520* \\
RAG-Random         & 0.729* & 0.728* & 0.460* & 0.545* \\
\rowcolor{lightblue} RAG-Nearest & 0.787 & 0.776 & 0.535 & 0.606 \\
\bottomrule
\end{tabular}}
  \end{minipage}
  \hfill
  \begin{minipage}{0.38\textwidth}
    \centering
        \caption{Dice score and HD95 computed between reference masks and masks predicted from generated CTs.}
    \label{tab:dice}
    \resizebox{\linewidth}{!}{
    \begin{tabular}{l|cc}
\toprule
\textbf{Model} & \textbf{DICE ↑} & \textbf{HD95 ↓} \\
\midrule
Real & 0.847 & 1.872 \\
\midrule
MAISI~\cite{guo2025maisi} & 0.792 & 2.811 \\
\midrule
RAG-Farthest & 0.697* & 5.447* \\
RAG-Random   & 0.710* & 5.391* \\
\rowcolor{lightblue} RAG-Nearest & 0.772 & 3.072  \\
\bottomrule
\end{tabular}}
  \end{minipage}
\end{table}

\noindent\textbf{Clinical Consistency}
Table~\ref{tab:ctnet} reports classification performance on synthetic CT volumes generated by the different methods.
For reference, CT-Net achieves an AUC of 0.824 when evaluated on real CT volumes, providing an upper-bound performance estimate.
MAISI is not included in this evaluation, as it conditions solely on segmentation masks and does not incorporate report-based semantic guidance.
Retrieval-augmented generation improves diagnostic performance with respect to text-only baselines, indicating better preservation of clinically meaningful patterns.
Semantically nearest retrieval achieves the strongest performance, highlighting the importance of semantic alignment between the report and the retrieved structural proxy.
Random and farthest retrieval degrade performance, confirming the sensitivity of clinical realism to retrieval quality.

\noindent\textbf{Spatial Controllability}
Table~\ref{tab:dice} reports Dice and HD95, measuring the degree to which each method adheres to the provided structural conditioning signal.
This evaluation is restricted to methods incorporating structural conditioning, as comparing text-only methods against a reference mask would only measure accidental overlap.
Specifically, for RAG variants, generated volumes are segmented and compared against the retrieved proxy, measuring spatial controllability as adherence to the provided scaffold, rather than anatomical correctness with respect to unknown ground-truth.
MAISI conditions directly on ground-truth masks, representing a structural upper bound by design rather than a directly comparable approach.
Real CT volumes are reported to establish a reference for the segmentation pipeline.
RAG-Nearest approaches MAISI's structural adherence while preserving semantic flexibility. 
Notably, a model trivially copying the proxy would score high here but degrade on Table~\ref{tab:ctnet}, confirming that the evaluation axes are complementary to assess generation quality.

\begin{figure*}[t]
    \centering
    \includegraphics[width=0.76\textwidth]{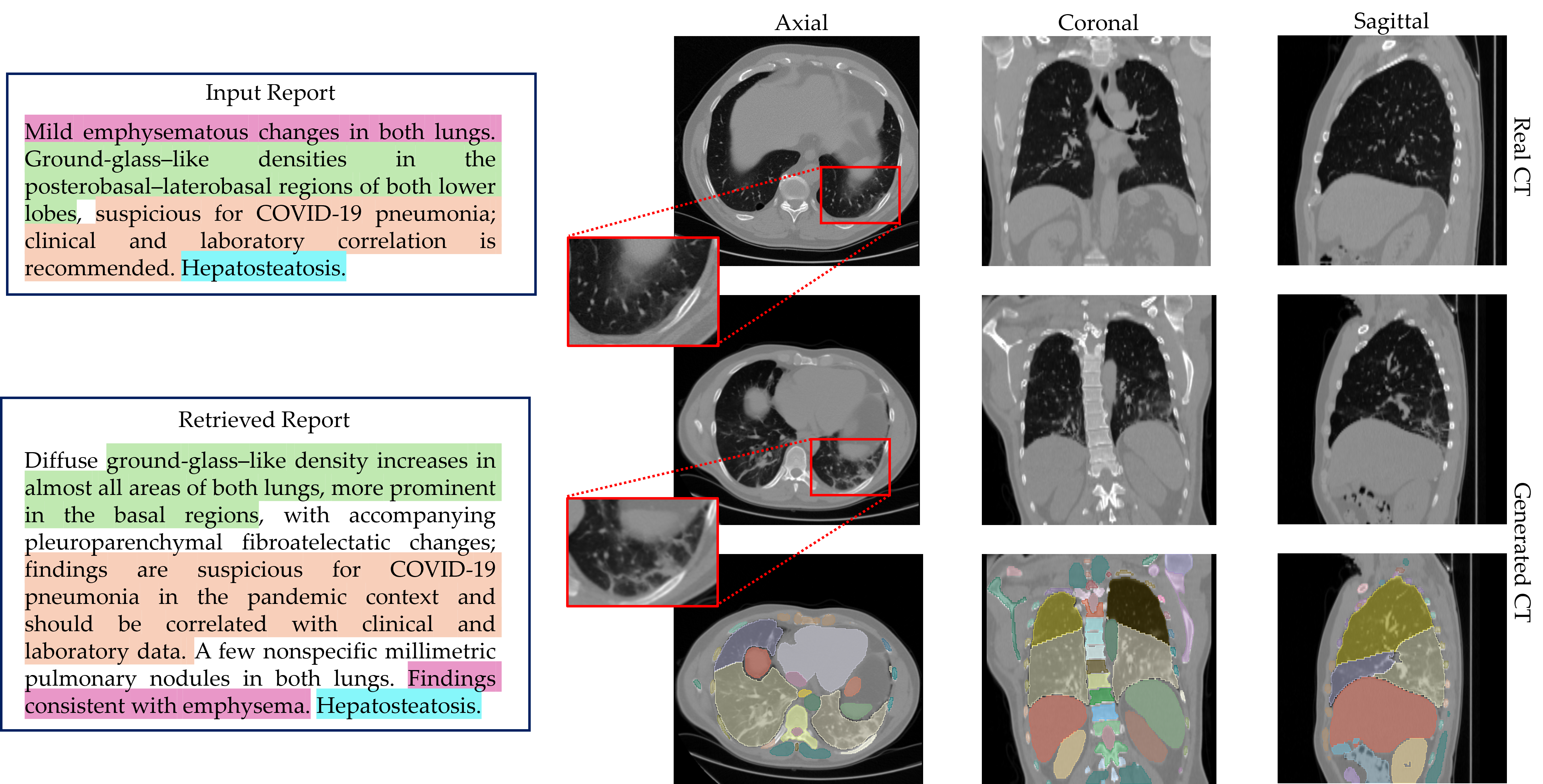}
    \caption{Qualitative example of retrieval-augmented generation. Left: input report and retrieved report, with overlapping clinical concepts highlighted. Right: real CT and the generated CT with and without the retrieved anatomical proxy $m$.}
    \label{fig:qualitative}
\end{figure*}

\noindent\textbf{Qualitative Comparison}
Figure~\ref{fig:qualitative} illustrates a representative example.
The retrieved case exhibits semantic overlap with the input report, and the generated CT adheres to the spatial prior provided by the proxy mask.
In particular, the retrieved anatomical scaffold constrains the global thoracic layout, reducing spatial ambiguity while allowing local variations consistent with the report.

\section{Conclusion}
\label{sec:conclusion}
We introduced a retrieval-augmented framework for Text-to-CT generation that bridges semantic and anatomical conditioning without requiring annotations at inference time. 
Anatomy is modeled as a retrievable proxy in a shared 3D vision-language embedding space, improving coherence while preserving report-driven variability. 
Experiments on CT-RATE demonstrate consistent gains in image fidelity, clinical consistency, and spatial controllability. Future work will investigate pathology-specific evaluation and longitudinal scenarios, leveraging temporally related priors to model disease progression.



\bibliographystyle{splncs04}
\bibliography{references}

\end{document}